\documentclass[10pt,journal,compsoc]{IEEEtran}

\ifCLASSOPTIONcompsoc
  \usepackage[nocompress]{cite}
\else
  \usepackage{cite}
\fi

\ifCLASSINFOpdf
\else
\fi

\usepackage{graphics,amssymb,amsmath,epsfig,color}
\usepackage{graphicx}
\usepackage{microtype}
\usepackage{graphicx}
\usepackage{booktabs} % for professional tables

\usepackage{amsmath} % boldsymbol
\usepackage{amssymb,bm} % mathbb
\usepackage{mathtools}% 
\usepackage{cases} %subnumb cases
\usepackage{amsfonts}
\usepackage{float}
\usepackage[algoruled,linesnumbered]{algorithm2e}
\usepackage{multirow}
\usepackage{tikz}
\usetikzlibrary{matrix,decorations.pathreplacing, calc, positioning}
\usepackage{subcaption,algorithmic}

\usepackage{hyperref}

\DeclareFontFamily{U}{mathx}{\hyphenchar\font45}
\DeclareFontShape{U}{mathx}{m}{n}{
	<5> <6> <7> <8> <9> <10>
	<10.95> <12> <14.4> <17.28> <20.74> <24.88>
	mathx10
}{}
\DeclareSymbolFont{mathx}{U}{mathx}{m}{n}
\DeclareFontSubstitution{U}{mathx}{m}{n}
\DeclareMathAccent{\widecheck}{0}{mathx}{"71}
\DeclareMathAccent{\wideparen}{0}{mathx}{"75}

\newcommand{\myvec} [1]{\bm{#1}} %vector
\newcommand{\mymat} [1]{\bm{#1}}%matrix
\newcommand{ \col}[1]{\text{range}\left( #1 \right) }  % column space of matrix
\newtheorem{theorem}{Theorem}[section]

\newtheorem{lemma}[theorem]{Lemma}

\begin{document}
\title{Generalized Canonical Correlation Analysis:\\ A Subspace Intersection Approach}

\author{Mikael~S\o rensen,
%	~\IEEEmembership{Member,~IEEE,}
        Charilaos~I.~Kanatsoulis,
%        ~\IEEEmembership{Fellow,~OSA,}
        and~Nicholas~D.~Sidiropoulos
%        ~\IEEEmembership{Life~Fellow,~IEEE}% <-this % stops a space
\IEEEcompsocitemizethanks{\IEEEcompsocthanksitem M. S\o rensen and N. D. Sidiropoulos are with the Department
of Electrical and Computer Engineering, University of Virginia, Charlottesville,
VA, 22904.\protect\\
E-mail: (ms8tz,nikos)@virginia.edu
\IEEEcompsocthanksitem C. I. Kanatsoulis is with the Department
of Electrical and Computer Engineering, University of Minnesota, Minneapolis,
MN, 55455.\protect\\% <-this % stops an unwanted space
E-mail: kanat003@umn.edu}
}

\IEEEtitleabstractindextext{%
\begin{abstract}
Generalized Canonical Correlation Analysis (GCCA) is an important tool that finds numerous applications in data mining, machine learning, and artificial intelligence. It aims at finding `common' random variables that are strongly correlated across multiple feature representations (views) of the same set of entities. CCA and to a lesser extent GCCA have been studied from the statistical and algorithmic points of view, but not as much from the standpoint of linear algebra. This paper offers a fresh algebraic perspective of GCCA based on a (bi-)linear generative model that naturally captures its essence. It is shown that from a linear algebra point of view, GCCA is tantamount to subspace intersection; and conditions under which the common subspace of the different views is identifiable are provided. A novel GCCA algorithm is proposed based on subspace intersection, which scales up to handle large GCCA tasks. Synthetic as well as real data experiments are provided to showcase the effectiveness of the proposed approach.
\end{abstract}

\begin{IEEEkeywords}
Canonical Correlation Analysis, Generalized Canonical Correlation Analysis, Subspace Intersection, Multiview Learning, Identifiability, Algebraic Algorithm.
\end{IEEEkeywords}}

\maketitle
\section{Introduction}
\label{intro}
Canonical Correlation Analysis (CCA) is a classical statistical tool for two-set  / two-view factor analysis \cite{Hotelling36,Golub95}. It aims at extracting a common latent structure of a set of entities observed in two different feature domains, which are usually referred as the `views' of the entities. For example, an English document and its French translation is an entity represented in two different language-views. CCA can be naturally extended to the multi-view case, where more than two views are available for processing. Then it is referred as generalized CCA (GCCA) or multi-view CCA (MCCA) \cite{Kettenring71}. CCA/GCCA can also be considered as an extension of principal component analysis (PCA) to the case where multiple views of the data are available. On the one hand PCA seeks for a feature representation that maximizes the variance explained, thus keeping the strong / principal feature components. On the other hand, CCA/GCCA extracts the common components between the views and ideally ignores even strong components that are not present in all the views.

(G)CCA is a powerful set of tools with diverse applications in machine learning \cite{Vinokourov03,Hardoon04,Dhillon11,Andrew13}, data mining \cite{multi_clust,rastogi2015multiview,fu2016efficient,Kanatsoulis19}, signal processing \cite{Via07,VanVaerenbergh13,arora2014multi,Vasquez14,Ibrahim19}, biomedical engineering \cite{Borga01,DeClercq06,Campi13}, heath care data analytics \cite{vasquez2017multi}, and genetics \cite{parkhomenko2009sparse,witten2009extensions}, among others.  

In the two view case, CCA can be optimally solved via generalized eigenvalue decomposition \cite{Golub95}. Furthermore, several algorithms exist that solve the CCA problem when big and high dimensional datasets are involved, and eigenvalue solutions are computationally prohibitive, e.g., \cite{lu2014large,ge2016efficient}. The multi-view scenario, on the other hand, is more complicated. There exist a number of different GCCA formulations, e.g., SUMCOR, MAXVAR, SUQUAR, etc; see \cite{carroll1968generalization,Kettenring71}, and the majority of them are not solvable in polynomial time. SUMCOR and MAXVAR are the most popular formulations and various algorithms have been developed for them, e.g., \cite{rastogi2015multiview,arora2014multi,vasquez2017multi,fu2016efficient,Kanatsoulis19}.

Although CCA and GCCA are well-known and broadly-used tools with a long history, there still exist intriguing questions and open challenges related to (G)CCA theory and practice. First, our understanding of CCA/GCCA from an algebraic perspective is limited. The majority of the literature focuses on the statistical interpretation of CCA, e.g., \cite{Hotelling36,Kettenring71,bach2005probabilistic}, where each view is considered as a set of random vector realizations, and/or on algorithmic aspects. Interpreting (G)CCA from an algebraic viewpoint is important, since in practice the matrix views involved in (G)CCA do not necessarily follow a statistical model. Second, identifiability of CCA/GCCA, i.e., conditions under which the common latent components can be recovered, has only been partially studied. An identifiability condition for CCA was derived in \cite{Ibrahim19}, but only for the two view case. Finally, there is limited analysis regarding the effect of multiple views compared to just using two views. Despite the rapid developments in data acquisition and cross-platform data availability, which enable leveraging multiple views of a given set of entities, researchers often work with just two views due to the more complicated nature of GCCA.

In this work we give answers to the above research questions. We show that from an algebraic point of view, GCCA amounts to {\em subspace intersection}, i.e., it computes the intersection of the subspaces of the given matrix views.  Furthermore, we provide improved and general conditions under which the common subspace between the views is identifiable. Our conditions show that having access to more views which share a common subspace benefits the identifiability of that subspace. {In addition, we propose a simple and effective subspace intersection algorithm for GCCA which works for any number of views greater than or equal to two. The algorithm is algebraic and it exploits knowledge of the desired rank (useful signal rank, i.e., the dimension of the dominant information-bearing `signal subspace') of the matrix views. We also develop a large-scale approximation algorithm which works for big and high-dimensional data, both dense and sparse. Extensive simulations with synthetically generated and real datasets showcase the effectiveness of our proposed framework.}

\textbf{Notation:} Vectors, matrices and subspaces are denoted by lower case boldface, upper case boldface and upper case italic respectively. The $r$-th column, transpose, Frobenius norm, rank, range and kernel of a matrix $\mymat{A}$ are denoted by $\myvec{a}_r,~\mymat{A}^T,~\lVert\mymat{A}\rVert_F,~\text{rank}(\mymat{A}),~\col{\mymat{A}},~\text{ker}(\mymat{A})$, respectively. The symbol $\oplus$ denotes the direct sum between two subspaces, $\bigcap$ is the subspace intersection operator, $\mathit{B}^{\perp}$ is the orthogonal complement of subspace $\mathit{B}$ and $\text{dim}{({\mathit{B}})}$ denotes the dimension of subspace $\mathit{B}$.

\section{GCCA data model}
We begin our discussion by reviewing the classical CCA formulation \cite{Golub95,Hardoon04}:
\begin{subequations}\label{eq:cca}
	\begin{align}
	\max_{\mymat{Q}^{(1)},\mymat{Q}^{(2)}}&~{\rm Tr}\left(\mymat{Q}^{(1)T}\mymat{X}^{(1)T} \mymat{X}^{(2)}\mymat{Q}^{(2)}\right)\\
	{\rm s.t.}&~\mymat{Q}^{(n)T}\mymat{X}^{(n)T}\mymat{X}^{(n)}\mymat{Q}^{(n)} = \mymat{ I}_R,~n=1,2, \label{eq:norm}
	\end{align}
\end{subequations}
where $\mymat{X}^{(n)}\in \mathbb{C}^{I \times K_n}$ is the $n$-th view containing $I$ entities measured in a $K_n$-dimensional feature space, $\mymat{Q}^{(n)}\in \mathbb{C}^{K_n\times R}$ is a matrix that projects $\mymat{X}^{(n)}$ onto a subspace of dimension $R$ and $\mymat{I}_R$ represents the $R\times R$ identity matrix. The previous optimization form can be equivalently written as:
\begin{subequations}\label{eq:cca2}
	\begin{align}
	\min_{\mymat{Q}^{(1)},\mymat{Q}^{(2)}}&~\lVert\mymat{X}^{(1)}\mymat{Q}^{(1)}- \mymat{X}^{(2)}\mymat{Q}^{(2)}\rVert_F\\
	{\rm s.t.}&~\mymat{Q}^{(n)T}\mymat{X}^{(n)T}\mymat{X}^{(n)}\mymat{Q}^{(n)} = \mymat{ I}_R,~n=1,2. \label{eq:norm2}
	\end{align}
\end{subequations}
Therefore in the noiseless case the optimal CCA solution gives:
\begin{subequations}\label{eq:cca3}
	\begin{align}
	&\mymat{X}^{(1)}\mymat{Q}^{(1)}= \mymat{X}^{(2)}\mymat{Q}^{(2)}\\
	{\rm s.t.}&~\mymat{Q}^{(n)T}\mymat{X}^{(n)T}\mymat{X}^{(n)}\mymat{Q}^{(n)} = \mymat{ I}_R,~n=1,2, \label{eq:norm3}
	\end{align}
\end{subequations}
Equivalently, when multiple views are considered, the solution in the noiseless case should satisfy:
\begin{subequations}\label{eq:gcca}
	\begin{align}
	&\mymat{X}^{(n_1)}\mymat{Q}^{(n_1)}= \mymat{X}^{({n_2})}\mymat{Q}^{({n_2})},~n_1\neq n_2\label{equal1}\\
	{\rm s.t.}&~\mymat{Q}^{(n)T}\mymat{X}^{(n)T}\mymat{X}^{(n)}\mymat{Q}^{(n)} = \mymat{ I}_R,~n,n_1,n_2\in\{1,\dots, N\} \label{eq:norm4}
	\end{align}
\end{subequations}
It is straightforward to see that if \eqref{equal1} holds then the range of the views $\mymat{X}^{(1)}, \ldots,   \mymat{X}^{(N)}$ share a common $R$-dimensional subspace and thus admit the following multi-view factorization:
\begin{equation}
\mymat{X}^{(n)}=    
[\mymat{M},    \mymat{C}^{(n)}]\mymat{S}^{(n)T}, \quad n \in \{1, \ldots, N\},
\label{eq:MFA_model}
\end{equation}
where  $\mymat{M}  \in \mathbb{C}^{I \times R}$ is a shared factor matrix, while  $\mymat{C}^{(n)} \in \mathbb{C}^{I \times L_n}$ and $\mymat{S}^{(n)} \in \mathbb{C}^{K_n \times (R+L_n)}$ are  individual matrices for all $n \in \{1, \ldots, N\}$. The goal of generalized CCA is to find the subspace $\text{range}(\mymat{M})$, observing  $\mymat{X}^{(1)}, \ldots,   \mymat{X}^{(N)}$. Note the factorization in \eqref{eq:MFA_model} also holds for the standard two-view CCA, in which $N=2$. 

As far as identifiability is concerned, \cite{Ibrahim19} studied the two-view CCA and proved that if the matrices $[\mymat{M}   ,    \mymat{C}^{(1)}, \mymat{C}^{(2)}]$, $\mymat{S}^{(1)}$ and $\mymat{S}^{(2)}$ have full column rank, then $\col{\mymat{M}}$ can be obtained via CCA. In this paper we move a step forward and provide an identifiability condition for the general case of GCCA. More precisely, using the range subspace intersection approach for GCCA, we propose  an identifiability condition that does not require any of the matrices in the set $\{[ \mymat{M}   ,    \mymat{C}^{(n_1)}, \mymat{C}^{(n_2)}]\}$ to have  full column rank.   

Our first observation is that since $\mymat{M}$ is a shared factor matrix, we can {without loss of generality (w.l.o.g.)}  assume that the matrices    $\{[\mymat{M}    , \mymat{C}^{(n)}]\}$   in   \eqref{eq:MFA_model} have full column rank (see Appendix \ref{app:prop} for detailed proof). Next, we assume for simplicity that the matrices $\{\mymat{S}^{(n)}\}$ in \eqref{eq:MFA_model} have full column rank. {This assumption is used for brevity, although it is not generally necessary for identifiability of the common subspace. In the case where $\{\mymat{S}^{(n)}\}$ matrices do not have full column rank, the analysis becomes more complicated and is reserved for future work.} The full column rank property of $\{[\mymat{M},\mymat{C}^{(n)}]\}$ and the full column rank assumption on $\{\mymat{S}^{(n)}\}$ 
{allow us to assume, w.l.o.g. that}\footnote{{Recall that if $\mathit{U}$ and $\mathit{V}$ are subspaces of the vector space $\mathit{W}$, then $\mathit{W}=\mathit{U}\oplus \mathit{V}$   if and only if  $\mathit{U} \cap \mathit{V}=\{ \myvec{0} \}$ and   $\mathit{W}=\mathit{U}+\mathit{V} $.  Equivalently,  $\mathit{W}=\mathit{U}\oplus \mathit{V}$ if and only if  for    any $ \myvec{w}  \in \mathit{W}$  there exists a unique vector  $ \myvec{u}  \in \mathit{U}$ and  a unique vector  $ \myvec{v}  \in \mathit{V}$ such  that  $\myvec{w}=\myvec{u}+\myvec{v}$.
}}
\begin{equation}
\text{range}(\mymat{X}^{(n)})=\text{range}(\mymat{M})\oplus\text{range}(\mymat{C}^{(n)}), \quad n \in \{1,\ldots, N\},
\label{eq:X_MCn}
\end{equation}
which implies that  
\begin{equation}
R \leq I-\max_{1 \leq n \leq N}L_n.
\label{eq:inequality1}
\end{equation}
Note that equation \eqref{eq:X_MCn} is a key point of our approach and follows naturally from the fact that matrix $[\mymat{M}, \mymat{C}^{(n)}]$  has full column rank and that the subspaces $\text{range}(\mymat{M})$ and $\text{range}( \mymat{C}^{(n)})$ are complementary (see Appendix \ref{app:prop} for details). 

\section{A range subspace intersection approach for generalized  CCA} 
\label{sec:RangeApproach}
In this section we provide a  range subspace intersection approach  for finding    $\col{\mymat{M}}  $ via the observed matrices $ \mymat{X}^{(1)}, \ldots,  \mymat{X}^{(N)}$.  The full column rank assumptions on the matrices $\{[\mymat{M}    , \mymat{C}^{(n)}]\}$  and $\{\mymat{S}^{(n)}\}$  in   \eqref{eq:MFA_model}  imply that 
\begin{align*}
&\text{range}(  \mymat{X}^{(n_1)}) \cap  \text{range}(  \mymat{X}^{(n_2)})
\nonumber \\&= \text{range}( [\mymat{M}  , \mymat{C}^{(n_1)}]) \cap  \text{range}( [\mymat{M}  ,  \mymat{C}^{(n_2)}])
\nonumber \\
&= \text{range}( \mymat{M}   )  + \left(    \text{range}(  \mymat{C}^{(n_1)}) \cap  \text{range}(  \mymat{C}^{(n_2)}) \right), \\
&\hspace{15em} 1 \leq n_1  < n_2 \leq N,
\end{align*}
where the last equality follows from \eqref{eq:X_MCn}. More generally, we have that 
\begin{align}
Y&:=\bigcap_{n=1}^N \text{range}(  \mymat{X}^{(n)})=
\text{range}(  \mymat{M}  )  +     \bigcap_{n=1}^N   \text{range}(  \mymat{C}^{(n)}) \nonumber \\
&=  \text{range}( \mymat{M} )  \oplus \left(    \bigcap_{n=1}^N   \text{range}(  \mymat{C}^{(n)})  \cap \text{range}( \mymat{M} )^{\perp} \right) \nonumber \\
&=  \text{range}( \mymat{M}  )  \oplus  C,
\label{eq:2}
\end{align}
where $C:=    \bigcap_{n=1}^N   \col{\mymat{C}^{(n)}}  \cap \text{range}(  \mymat{M} )^{\perp}$. 
Observe that $\text{dim}( Y)=R$ implies that  $Y=\text{range}( \mymat{M}  )$. To put it differently, if the dimension of the subspace spanned by a basis for $Y$ is $R$-dimensional, then $\text{range}( \mymat{M}  )$ can be uniquely determined from  $Y$.  

What remains to be answered is how to determine the dimension of $Y$. Consider a nonzero vector $\myvec{z} \in \mathbb{C}^{I}$.  Since the  columns of  $\mymat{X}^{(n)} \in \mathbb{C}^{I \times (R+L_n)}$ form a basis for  $\text{range}(  \mymat{X}^{(n)}) $, we know that 
$\myvec{z} \in Y$ if and only if there exist nonzero vectors $\pmb{q}^{(1)}\in \mathbb{C}^{R+L_1}, \ldots,  \pmb{q}^{(N)}\in \mathbb{C}^{R+L_N}$ such that   
\begin{equation}
\label{eq:z_vector}
\myvec{z}=  \mymat{X}^{(1)} \pmb{q}^{(1)}=\cdots =  \mymat{X}^{(N)}\pmb{q}^{(N)}.
\end{equation}
Define   $\pmb{q}=[\pmb{q}^{(1)T}, \ldots, \pmb{q}^{(N)T}]^T \in \mathbb{C}^{(NR+\sum_{n=1}^NL_n)}$.  Then a vector  $\pmb{q}$ with property  \eqref{eq:z_vector} can be obtained by solving the system of homogenous linear equations
\begin{equation}
\left[\mymat{0}_{I,\alpha_{n_1}},  \mymat{X}^{(n_1)},\mymat{0}_{I,\beta_{n_1,n_2}}, -\mymat{X}^{(n_2)},\mymat{0}_{I,\omega_{n_2}}  \right]
\pmb{q}=\mymat{0}_{I},
\label{eq:homogen}
\end{equation}
for $1 \leq n_1 < n_2 \leq N$, where \\
$\alpha_{n_1}=(n_1-1)R+\sum_{i=1}^{n_1-1}L_i$,\\
$\beta_{n_1,n_2}=(n_2-n_1-1)R+\sum_{i=n_2+1}^{n_2-1}L_i$,\\
$\omega_{n_2}=(N-n_2)R+\sum_{i=n_2+1}^NL_i$.

We can now conclude that if the subspace 
{\begin{align}
	&Z^{(N)}:=
	%&{\bigcap_{1 \leq n_1 < n_2 \leq N} \text{ker}([ \mymat{0}_{I,\alpha_{n_1}},  \mymat{X}^{(n_1)},\mymat{0}_{I,\beta_{n_1,n_2}}, -\mymat{X}^{(n_2)},\mymat{0}_{I,\omega_{n_2}}  ])}\nonumber \\
	{\bigcap_{1 \leq n_1 < n_2 \leq N} \text{ker}([ \mymat{0},  \mymat{X}^{(n_1)},\mymat{0}, -\mymat{X}^{(n_2)},\mymat{0}  ])}\nonumber \\
	&={\bigcap_{1\leq n_1 < n_2\leq N} \text{ker}([  \mymat{0},  \mymat{M},    \mymat{C}^{(n_1)},\mymat{0},- \mymat{M}, -\mymat{C}^{(n_2)} ,\mymat{0}}])
	\label{eq:Z}
	\end{align}}
\noindent is $R$-dimensional, i.e., there exist only $R$ linearly independent vectors $\pmb{q}^{(1)}, \ldots,  \pmb{q}^{(N)}$ in the range of $Z^{(N)}$, then $ Y$  is also $R$-dimensional and  $Y=\text{range}( \mymat{M})$. Note that the dimensions of the zero matrices in \eqref{eq:Z} are as in \eqref{eq:homogen}, but have been omitted due to space limitations.

Therefore the  dimension of $ Y$  can  be expressed in terms of the factor matrices $\mymat{M},\{\mymat{C}^{(n)}\}_{n=1}^N$. For example, when $N=3$, the dimension of $ Y$ is equal to the dimension of the kernel of the following matrix (the same reasoning holds true when $N>3$):
\begin{align}\label{mat:kernel}
\left[\begin{array}{cccccc}
\mymat{M}   &     \mymat{C}^{(1)}&  -\mymat{M}   &     -\mymat{C}^{(2)}& \mymat{0}    &    \mymat{0} \\
\mymat{M}   &    \mymat{C}^{(1)}&   \mymat{0}   &     \mymat{0} &-\mymat{M}   &    -\mymat{C}^{(3)}\\
\mymat{0}   &    \mymat{0} &   \mymat{M}   &      \mymat{C}^{(2)}& -\mymat{M}   &    -\mymat{C}^{(3)}
\end{array}  \right].
\end{align}

\section{Identifiability condition for generalized CCA}
\label{sec:ID_GCCA}
Since $\text{range}( \mymat{M} ) \subseteq  Y$, the minimal dimension of  the subspace $Y$ given by  \eqref{eq:2} is $R$.  This also means that if the subspace  
$  Z^{(N)}$ given by   \eqref{eq:Z} is  $R$-dimensional, then $C=\{\myvec{0}\}$.\footnote{{Recall that the dimension of a direct sum is the sum of the dimensions of its summands. This fact also explains that if $Y$ is   $R$-dimensional, then $C=\{\myvec{0}\}$.}} This fact leads   to  the   common  subspace identifiability condition presented in Theorem   \ref{prop:1}:
\begin{theorem}
	\label{prop:1}
	Consider the multi-view   factorization of $\mymat{X}^{(1)},\ldots,  \mymat{X}^{(N)}$  given by   \eqref{eq:MFA_model}. If
	\begin{equation}
	\left\{ \begin{aligned}
	&Z^{(N)} \text{ is $R$-dimensional},\\ 
	&\mymat{S}^{(1)}, \ldots, \mymat{S}^{(N)} \text{  have full column rank}, 
	\end{aligned}
	\right.
	\label{eq:common_subspace_condition}
	\end{equation}
	then  $\text{dim}\left(\bigcap_{n=1}^N \text{range}(  \mymat{X}^{(n)}) \right)=R$ \\and ~$\bigcap_{n=1}^N \text{range}(  \mymat{X}^{(n)})=\text{range}( \mymat{M} )$.
\end{theorem}

Note that  checking the dimension of  $Z^{(N)}$ in  \eqref{eq:common_subspace_condition} can be cumbersome and it is not obvious how it is related to the factor matrices $\mymat{M}, \mymat{C}^{(1)}, \ldots,   \mymat{C}^{(N)}$ in \eqref{eq:MFA_model}. In order to obtain a simpler condition for the recovery of $\text{range}( \mymat{M})$ via $  \mymat{X}^{(1)}, \ldots,  \mymat{X}^{(N)}$, the following identity will be used \cite{Tiann02}:
\begin{align}
\label{eq:A1}
&\text{dim}\left(\bigcap_{n=1}^N \text{range}(  \mymat{X}^{(n)}) \right)\nonumber\\&=\sum_{n=1}^N\text{rank}( \mymat{X}^{(n)})-\text{rank}(\pmb{\Gamma}(  \mymat{X}^{(1)}, \ldots,  \mymat{X}^{(N)} )),
\end{align}
where the matrix $\pmb{\Gamma}(  \mymat{X}^{(1)}, \ldots,  \mymat{X}^{(N)} )$ is defined as follows:
\begin{equation*}
\pmb{\Gamma}(  \mymat{X}^{(1)}, \ldots,  \mymat{X}^{(N)} )=
\left[\begin{array}{cccc}
\mymat{X}^{(1)}&-   \mymat{X}^{(2)}& &\\
\vdots & & \ddots&\\
\mymat{X}^{(1)} &&&  - \mymat{X}^{(N)} \\
\end{array}\right] ,  
\end{equation*}
in which $  \mymat{X}^{(1)}, \ldots,  \mymat{X}^{(N)}$ are matrices of conformable sizes.  Relation  \eqref{eq:A1} together with the full column rank assumptions on $\mymat{S}^{(1)}, \ldots, \mymat{S}^{(N)} $ imply that 
\begin{align}
&\text{dim}\left(\bigcap_{n=1}^N \text{range}(  \mymat{X}^{(n)}) \right)=\nonumber\\&\sum_{n=1}^N\text{rank}( \mymat{X}^{(n)})-\text{rank}(\pmb{\Gamma}(  \mymat{X}^{(1)}, \ldots,  \mymat{X}^{(N)} ) )=\nonumber \\
&\sum_{n=1}^N\text{rank}([ \mymat{M} ,  \mymat{C}^{(n)} ] )-\text{rank}(\pmb{\Gamma}( [ \mymat{M} ,  \mymat{C}^{(1)} ], \ldots,  [ \mymat{M} , \mymat{C}^{(N)} ] ) ).
\label{eq:A2}
\end{align}
The full column rank property of
the matrices $ [ \mymat{M} ,  \mymat{C}^{(1)} ], \ldots,  [ \mymat{M}   ,  \mymat{C}^{(N)} ] $ in turn  imply that relation  \eqref{eq:A2} simplifies to  
\begin{align}
&\text{dim}\left(\bigcap_{n=1}^N \text{range}(  \mymat{X}^{(n)}) \right) = \nonumber \\ 
&NR+ \sum_{n=1}^N L_n % \nonumber \\
-\text{rank}(\pmb{\Gamma}( [ \mymat{M} ,  \mymat{C}^{(1)} ], \ldots,  [ \mymat{M} , \mymat{C}^{(N)} ] ) )\nonumber\\
& NR+ \sum_{n=1}^N L_n - \text{rank}\left(\pmb{\Gamma}^{(N)}\right),
\label{eq:A3}
\end{align}
where    the matrix $\pmb{\Gamma}^{(N)}\in \mathbb{C}^{(N-1)I \times ((N-1)R+\sum_{n=1}^NL_n)}$ in  \eqref{eq:A3} is given by
\begin{equation}
\pmb{\Gamma}^{(N)}=
\left[\begin{array}{ccccc}
\mymat{C}^{(1)}&- \mymat{M}&   - \mymat{C}^{(2)} &&\\
\vdots & &\ddots&\ddots&\\
\mymat{C}^{(1)}&&&- \mymat{M}&  - \mymat{C}^{(N)} \\
\end{array}\right] . 
\label{eq:GammaN}
\end{equation}
From  \eqref{eq:A3} it is clear that if   $\pmb{\Gamma}^{(N)}$    has full column rank, then  $\text{dim}\left(\bigcap_{n=1}^N \text{range}(  \mymat{X}^{(n)}) \right)=R$ and   $\bigcap_{n=1}^N \text{range}(  \mymat{X}^{(n)})=\text{range}(\mymat{M})$. Theorem \ref{prop:2}  below  summarizes the  simplified version of the common  subspace identifiability condition in Theorem  \ref{prop:1}.

\begin{theorem}
	\label{prop:2}
	Consider the multi-set   factorization of $\mymat{X}^{(1)},\ldots,  \mymat{X}^{(N)}$  given by   \eqref{eq:MFA_model}.  If
	\begin{equation}
	\left\{ \begin{aligned}
	&\pmb{\Gamma}^{(N)} \text{  has full column rank}, \\ 
	&\mymat{S}^{(1)}, \ldots, \mymat{S}^{(N)} \text{  have full column rank}, 
	\end{aligned}
	\right.
	\label{eq:common_subspace_condition2}
	\end{equation}
	then  $\text{dim}\left(\bigcap_{n=1}^N \col{\mymat{X}^{(n)}} \right)=R$\\ and ~$\bigcap_{n=1}^N \col{\mymat{X}^{(n)}}=\col {\mymat{M}}$.  
\end{theorem}

Note that if $\pmb{\Gamma}^{(N)}$ has full column rank, then the matrices $ [ \mymat{M} , \mymat{C}^{(1)} ], \ldots,  [ \mymat{M} ,  \mymat{C}^{(N)} ] $ must  also  have full column rank. 

Regarding necessary conditions so that \eqref{eq:common_subspace_condition2} is satisfied, we have:
\begin{equation}
\left\{ \begin{aligned}
&R+\frac{1}{N-1}\sum_{n=1}^NL_n\leq I,\\
&R+L_n \leq K_n  , \quad n \in \{1, \ldots, N \}. 
\end{aligned}
\right.
\label{eq:common_subspace_condition3}
\end{equation}

Now we consider the case where $\mymat{M},\mymat{C}^{(n)},\mymat{S}^{(n)}, n=1,\dots,N$ are generic. Then $R+L_n \leq K_n$ is also sufficient for $\mymat{S}^{(n)}$ to have full column rank. In order to study the rank of $\pmb{\Gamma}^{(N)}$, when $\mymat{M},\mymat{C}^{(n)}, n=1,\dots,N$ are generic, we will make use of the following lemma:
\begin{lemma}
	\label{lem:analytic}
	\cite{GunningBook} 
	Let   $f: \mathbb{C}^n \to \mathbb{C}$ be an analytic function. If there exists an element  $\myvec{x} \in \mathbb{C}^n $ such that $f \left( \myvec{x} \right) \neq 0$, then the set $\{\, \myvec{x} \, | \,  f\left(\myvec{x}\right)=0 \, \}$ is of Lebesgue measure zero.  
\end{lemma}

Recall that an $m \times n$ matrix  has full column rank $n$ if it  has  a non-vanishing $n \times n$ minor. Since a minor is an analytic  function  (namely, it is a  multivariate polynomial),      Lemma \ref{lem:analytic} can  now be used to verify whether the matrices in   \eqref{eq:common_subspace_condition2}  generically have full column rank. As a result, $R+\frac{1}{N-1}\sum_{n=1}^NL_n\leq I$ is generically sufficient for $\pmb{\Gamma}^{(N)}$ to have full column rank, if there exists a single example of $\mymat{M},\mymat{C}^{(n)}, n=1,\dots,N$ with det$\left(\pmb{\Gamma}^{(N)}\right)\neq 0$ for each tuple  $\{R, L_1, \ldots, L_N\}$. An illustrative derivation of such an example for a special but interesting choice of  $\{R, L_1, \ldots, L_N\}$ is provided in the supplementary material.
Numerical experiments indeed suggest that the conditions in \eqref{eq:common_subspace_condition3} are generically sufficient for generalized CCA identifiability.

\section{Further Discussion and Insights}
In this section we discuss the effect of processing more views (i.e., $N > 2$) using GCCA compared to the more commonly used CCA model in which $N=2$.

First we note that when $N=2$, condition  \eqref{eq:common_subspace_condition2} boils down to the standard two-view CCA identifiability condition obtained in  \cite{Ibrahim19}. However, when  $N>2$ condition \eqref{eq:common_subspace_condition2} yields relaxed identifiability. For example, consider the case where $\mymat{M},\mymat{C}^{(n)},\mymat{S}^{(n)}, n=1,\dots,N$ are generic and $R=L_n=100,~n=1,\dots,N$. Then condition \eqref{eq:common_subspace_condition3} for two-view CCA requires $I\geq R+L_1+L_2=300$, in order to recover $\col{\mymat{M}}$. When $N=3$, however, the condition in \eqref{eq:common_subspace_condition3} is relaxed to $I\geq R+\frac{1}{2}(L_1+L_2+L_3)=250$. Furthermore, in the latter case none of   the matrices $[ \mymat{M}   ,    \mymat{C}^{(1)}, \mymat{C}^{(2)}]$, $[ \mymat{M}   ,    \mymat{C}^{(1)}, \mymat{C}^{(3)}]$  and $[ \mymat{M}   ,    \mymat{C}^{(2)}, \mymat{C}^{(3)}]$ are  required to have full column rank, which is necessary in the two-view case. The identifiability condition for GCCA can be further relaxed by increasing $N$. In the previous example, when $N=5$ the condition in \eqref{eq:common_subspace_condition3} reduces to $I\geq 225$ and as $N\to \infty$ to $I\geq R+L_n=200$, which is also a necessary condition to identify $\col{\mymat{M}}$.

{Another important difference between the two-view and multi-view CCA is that when $N=2,~\col{\mymat{C}^{(1)}} \cap \col{\mymat{C}^{(2)}} = \{\myvec{0} \} $ is a necessary  identifiability condition. On the contrary, for $N>2$ it is possible that   $\col{\mymat{C}^{(m)}} \cap \col{\mymat{C}^{(n)}}\neq \{\myvec{0} \} $ for some $m \neq n$, i.e., some views are allowed to share common subspaces not included in $\col{\mymat{M}}$}.

To further elaborate on the identifiability properties of CCA ($N=2$) and GCCA  ($N>2$), consider the case where $R$ is fixed. Condition \eqref{eq:common_subspace_condition3} implies that 
\begin{equation}
\label{eq:GenericGCCA_3}
\sum_{n=1}^NL_n\leq  (N-1)(I-R)
\end{equation}
is necessary for condition  \eqref{eq:common_subspace_condition2} to be satisfied. If additionally $L_1=\cdots=L_N$, then \eqref{eq:GenericGCCA_3} reduces to: 
\begin{equation}
\label{eq:GenericGCCA_4}
L_n\leq   \frac{N-1}{N}(I-R),
\end{equation}
which yields the following relation, when $N=2$:
\begin{equation}
\label{eq:GenericGCCA_5}
L_n\leq   \frac{1}{2}(I-R).
\end{equation}
From \eqref{eq:GenericGCCA_4}, we can also infer that in the asymptotic case where $N \to \infty$, \eqref{eq:GenericGCCA_4} reduces to:
\begin{equation}
\label{eq:GenericGCCA_6}
L_n\leq   I-R.
\end{equation}
Comparing \eqref{eq:GenericGCCA_5} with \eqref{eq:GenericGCCA_6}, we conclude that in the balanced case where $L_1=\cdots=L_N$, GCCA can at most relax the CCA bound on $L_n$ by a factor two.

Let us now consider the balanced case ($L_1=\cdots=L_N$) where $L_n$ is fixed, while $R$ is varying. Relation  \eqref{eq:GenericGCCA_5} implies that CCA with $N=2$ views is able to recover the common subspace $\col{\mymat{M}}$ only if 
\begin{equation}
\label{eq:GenericGCCA_7}
R\leq I-2L_n.
\end{equation}
In other words, $L_n<\frac{I}{2}$  is  a necessary   recovery condition for CCA. On the contrary, employing more views ($N>2$), allows GCCA to recover the common subspace $\col{\mymat{M}}$, even if $L_n\geq \frac{I}{2}$. For instance, if $I=200$,  $N=5$ and  $L_1=\cdots=L_N=100$, then it can be verified that condition  \eqref{eq:common_subspace_condition3} is satisfied as a long as $R\leq 75$, regardless of the fact that  $L_n=\frac{I}{2}$. Furthermore, as $N \to \infty$ we get from \eqref{eq:GenericGCCA_6} that:
\begin{equation}
\label{eq:GenericGCCA_8}
R\leq   I-L_n
\end{equation}
is necessary to satisfy condition  \eqref{eq:common_subspace_condition2} in Theorem \ref{prop:2}. Comparing \eqref{eq:GenericGCCA_7} with \eqref{eq:GenericGCCA_8}, we conclude that when $L_1=\cdots=L_N$ and  $L_n<\frac{I}{2}$, GCCA can at most relax the bound on $R$ by a factor of two. Moreover, when $L_n\geq \frac{I}{2}$, GCCA can still ensure the  recovery of $\col{\mymat{M}}$ while this is \emph{never} possible when $N=2$.

\section{Algorithmic framework}
\label{sec:Algorithms}
In this section we develop an algebraic algorithm that uses a range subspace intersection approach to tackle the GCCA problem. The algorithm follows the previous analysis and can be viewed as a constructive interpretation of Theorem \ref{prop:1}. It can be described in 3 steps:

\noindent\textbf{step 1:} We compute $\mymat{U}_n \in \mathbb{C}^{I \times (R+L_n)}$ and $\mymat{V}_n \in \mathbb{C}^{K_n \times (R+L_n)}$ whose columns form an orthonormal basis for $\col{\mymat{X}^{(n)}}$ and $\col{\mymat{X}^{(n)T}}$, $n \in \{1, \ldots, N\}$ respectively. \\
\textbf{step 2:} This step forms a matrix {$\bm \Theta\in\mathbb{C}^{NI\times(NR+\sum_{n=1}^NL_n)}$} and computes a basis for it's nullspace, represented by matrix $\bm \Phi$. An example of matrix $\bm \Theta$ when $N=3$ can be found in \eqref{mat:kernel}. The columns of $\bm \Phi$ form an orthonormal basis for $Z^{(N)}$ in \eqref{eq:Z}. Note that $\bm\Phi=[   \bm\Phi^{(1)T} ,  \ldots,      \bm\Phi^{(N)T} ]^T$, with $\bm\Phi^{(n)}  \in \mathbb{C}^{(R+L_n) \times R}$. Then the matrices $\mymat{Q}^{(n)},~n=1,\dots,N$ that project each view $\mymat{X}^{(n)}$ to the common subspace $\col{\mymat{M}}$ can be obtained as $\mymat{Q}^{(n)}=\bm V_{n}\bm\Phi^{(n)}$. \\
\textbf{step 3:} We build matrix $\bm G=[ \mymat{U}^{(1)}   \bm\Phi^{(1)}, \ldots,   \mymat{U}^{(N)}   \bm\Phi^{(N)} ]$. Finally we compute matrix $\mymat{M}$ whose columns form an orthonormal basis for $\col{\bm G}$, which is the common subspace. The detailed steps can be found in Algorithm \ref{algo:RACING}.
\begin{algorithm}[tb]
	\caption{\texttt{RAnge subspaCe INtersection for Gcca (RACING)}}
	\label{algo:RACING}
	\begin{algorithmic}
		\STATE {\bfseries Input:} $\{\mymat{X}^{(n)},L_n\}_{n=1}^N,~R $.
		\STATE {\bfseries Output:} $\mymat{M},\bm Q^{(n)}$.
		\STATE {\bfseries step 1:}
		\FOR{$n=1$ {\bfseries to} $N$}
		\STATE $\mymat{U}_n \mymat{\Sigma}_n\bm V_n^T \leftarrow {\rm svds}\left( \mymat{X}^{(n)},'R+L_n' \right);$
		\ENDFOR
		\STATE {\bfseries step 2:}
		\STATE $\bm \Theta=[\cdot];$
		\FOR{$n_1=1$ {\bfseries to} $N-1$}
		\FOR{$n_2=n_1+1$ {\bfseries to} $N$}
		\STATE $k=(n_1-1)R+\sum_{1}^{n_1-1}L_n,~l=(n_2-n_1-1)R+\sum_{1}^{n_2-n_1-1}L_n,~m=(N-n_2)R+\sum_{n_2}^{N}L_n;$
		\STATE $\bm\theta_{n_1n_2}=[\bm 0_{I\times k},\bm U_{n_1},\bm 0_{I\times l},-U_{n_2},\bm 0_{I\times m}];$
		\STATE $\bm \Theta=[\bm\Theta;\bm\theta_{n_1n_2}];$
		\ENDFOR
		\ENDFOR
		\STATE $\bm U_{\theta}\bm \Sigma_{\theta}\bm V_{\theta}^T \leftarrow {\rm svd}\left( \bm \Theta,'{\rm econ}' \right);$
		\STATE $\bm \Phi=\bm V_{\theta}(:,\text{end}-R+1:\text{end});$
		\STATE {\bfseries step 3:}
		\STATE $\bm G=[\cdot];$
		\FOR{$n=1$ {\bfseries to} $N$}
		\STATE $k=(n-1)R+\sum_{1}^{n-1}L_n,~l=nR+\sum_{1}^{n}L_n;$
		\STATE $\bm G=[\bm G,\bm U_n\bm\Phi(k+1:l,:)];$
		\STATE $\mymat{Q}^{(n)}=\bm V_{n}\bm\Phi(k+1:l,:);$
		\ENDFOR
		\STATE $\bm U_{m}\bm \Sigma_{m}\bm V_{m}^T \leftarrow {\rm svds}\left( \bm G,'R' \right);$
		\STATE $\bm M=\bm U_{m};$
	\end{algorithmic}
\end{algorithm}

In terms of computational complexity and memory requirements, the main bottleneck of the proposed algorithm lies in step 1 and step 2. In particular, the dimension of the columnspace of each view $\bm X^{(n)}$ is usually large which makes the SVD computation in step 1 and step 2 very intensive. To be more precise, the number of flops required to perform the SVD in step 1 and step 2 are $\mathcal{O}(IK_n\min(I,K_n))$ and $\mathcal{O}(N(N-1)I(NR+\sum_{1}^{N}L_n)^2)$. To overcome this issue for high dimensional data we propose to choose $L_n$ such that $R+L_n$ represents the {useful signal rank, i.e., the dimension of the signal subspace,} which is small enough for every $n\in\{1,\dots, N\}$. Then Lanczos-type iterative algorithms \cite{larsen1998lanczos} can be used to compute the truncated SVD that significantly reduce the computational complexity, especially when sparse data are involved and $R+L_n\ll I$. For example, choosing $R+L_n$ to be in the order of $500$ will allow the proposed algorithm to work for very large and high-dimensional data.  
\section {Experiments}\label{sec:num}
In this section we demonstrate the performance of the proposed algorithmic framework and showcase it's effectiveness in synthetic- and real-data experiments. All simulations are implemented in Matlab and are executed on a Linux server comprising 32 cores at 2GHz and 128GB RAM.

\subsection{Synthetic-Data Experiments}
First we test the proposed framework using experiments with synthetically generated data.
The multiple views are generated according to equation \eqref{eq:MFA_model}. We assume that the views share a common latent factor ${\bm M} \in\mathbb{R}^{I\times R}$ with entries randomly and independently drawn from a zero-mean unit-variance Gaussian distribution. The individual matrices $\mymat{C}^{(n)} \in \mathbb{C}^{I \times L_n}$ and $\mymat{S}^{(n)} \in \mathbb{C}^{K_n \times (R+L_n)}$ are also generated with entries independently drawn from a zero-mean unit-variance Gaussian distribution and for simplicity we set $L_n=L$ and $K_n=K=L+R$ for every $n\in\{1,\dots, N\}$.

We test the algorithm in a noisy setup. To be more specific, $\bm X^{(n)},~n=1,\dots, N$ are generated according to the model in \eqref{eq:MFA_model}, as previouly described. However, instead of $\bm X^{(n)}$ we observe $\bm B^{(n)},~n=1,\dots, N$ which are generated as:
$
\bm B^{(n)}=\bm X^{(n)}+\bm W^{(n)},~~~~n\in\{1,\dots,N\},
$
where $\bm W^{(n)}$ is additive white Gaussian noise. Note that in the noiseless case the proposed algorithm is able to recover the common subspace exactly, for every tested scenario that satisfies the conditions in \eqref{eq:common_subspace_condition3}. For baselines we use the exact solution of \texttt{MAXVAR} formulation, computed via eigenvalue decomposition and \texttt{CSR-BCD}, which solves the SUMCOR formulation, using a change of variables and a block coordinate descent (BCD) approach \cite{fu2016efficient}. 
To evaluate the performance, we observe the maximum angle between the generated common subspace and the estimated one as defined in \cite{wedin1983angles,golub2012matrix}, i.e.,
\begin{equation}\label{angle}
\texttt{angle}\left(\bm M_1,\hat{{\bm M_1}}\right)=\lVert \bm P_1-\hat{\bm P_1}\rVert_2,
\end{equation}
where ${\bm P_1}(\hat{\bm P_1})$ is the orthogonal projection onto $\bm M_1(\hat{{\bm M_1}})$.

In the experiments we consider $N=6$ different views, that share a common subspace of dimension $R=50$. Two scenarios are generated: In the first each view consists of $I=2000$ rows, and $L=1000$ that leads to $K=1050$ columns for each view, whereas in the second $I=2000$, $Q=500$ that leads to $K=550$ columns for each view. We test the algorithmic performance for different levels of \emph{signal-to-noise-ratio} (SNR), which is defined as:
\begin{equation*}
SNR=20\log\frac{\sum_{n=1}^{N}\lVert\bm X^{(n)}\rVert_F}{\sum_{n=1}^{N}\lVert\bm W^{(n)}\rVert_F}.
\end{equation*} 

\begin{figure}
	\centering
	\begin{subfigure}[b]{0.23\textwidth}
		\includegraphics[width=\textwidth]{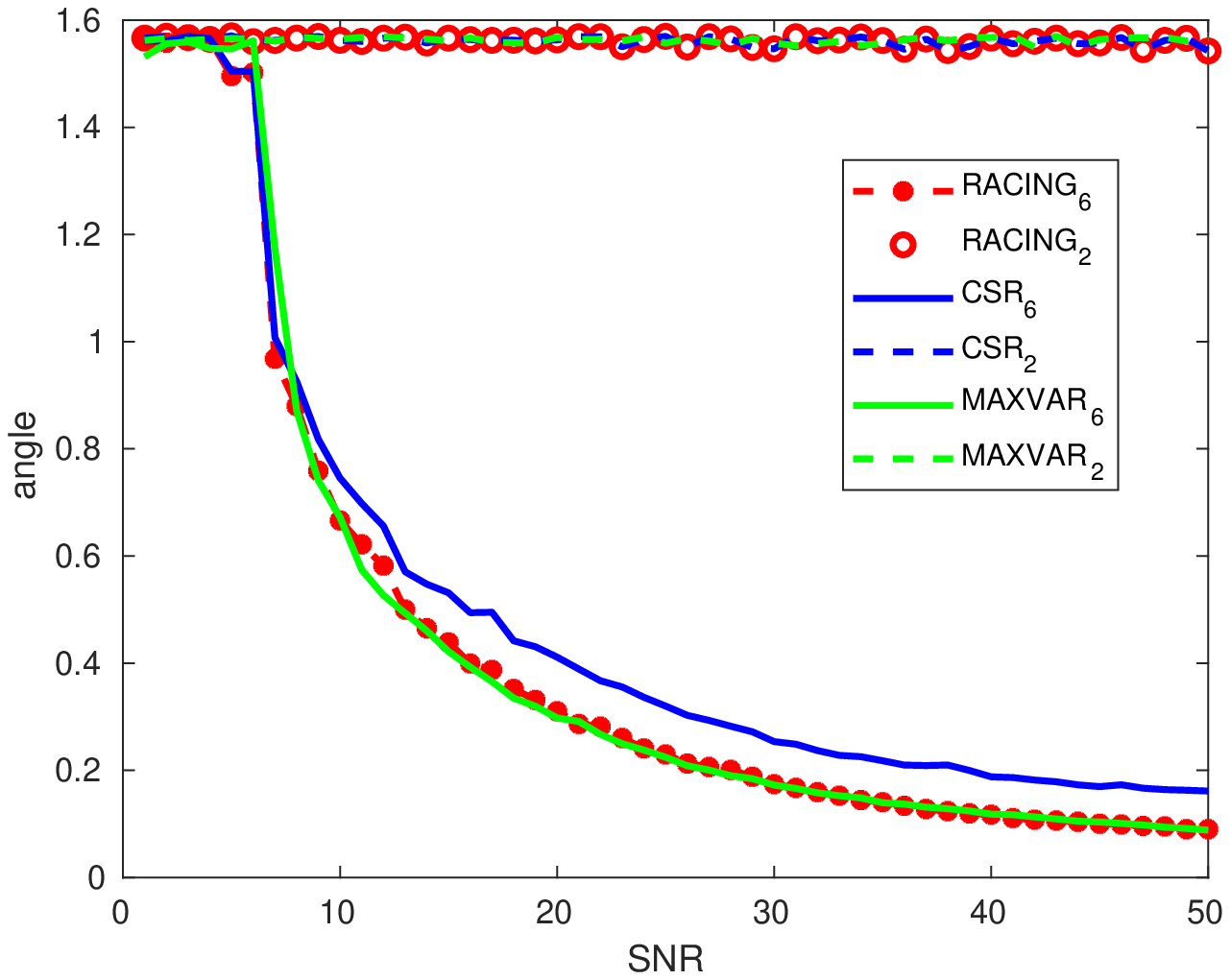}
		\caption{First scenario}
		\label{fig:first}
	\end{subfigure}
	%	~ %add desired spacing between images, e. g. ~, \quad, \qquad, \hfill etc. 
	%	%(or a blank line to force the subfigure onto a new line)
	\begin{subfigure}[b]{0.23\textwidth}
		\includegraphics[width=\textwidth]{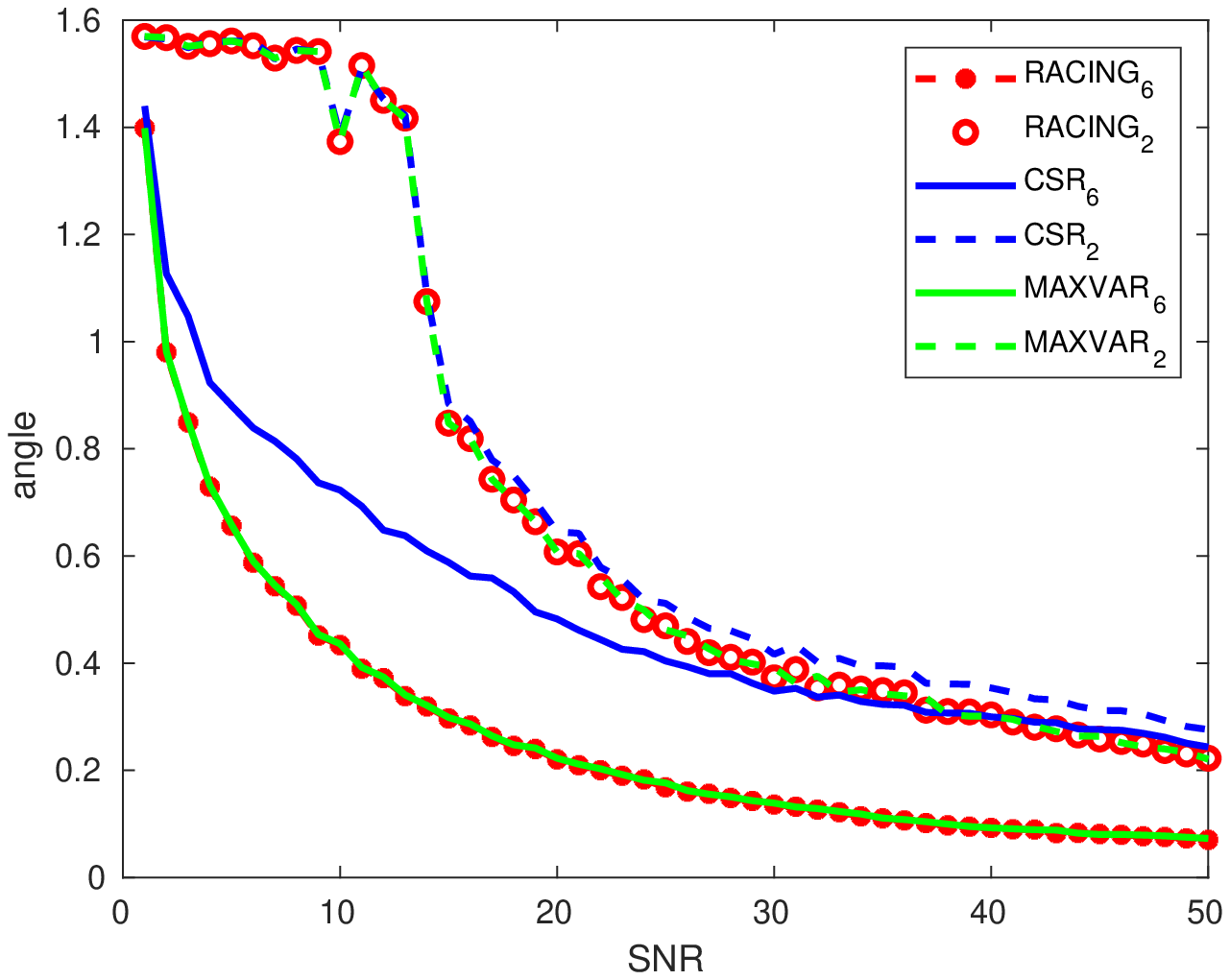}
		\caption{Second scenario}
		\label{fig:second}
	\end{subfigure}
	\caption{Angle between true and recovered subspace.}
	\label{fig:synthetic}
\end{figure}

Fig.~\ref{fig:synthetic}(a) shows the performance of the proposed \texttt{RACING} along the baselines for different levels of SNR in the first scenario. Note that each algorithm is implemented to utilize either all 6 views to identify the common space, or the first 2 views, denoted by the subscript next to the name of the algorithm. We observe that the proposed algorithm is able to identify the common subspace for a wide range of SNRs, when 6 views are utilized. On the contrary all algorithms fail to identify the correct subspace, when only 2 views are employed. Note that the identifiability condition in \eqref{eq:common_subspace_condition3} yields $50+\frac{N}{N-1}1000\leq 2000$, which is satisfied for $N=6$ but fails when $N=2$. 

In the second scenario we reduce the dimension of the columnspace of each view to $K=550$. In this case the identifiability condition in \eqref{eq:common_subspace_condition3} yields $50+\frac{N}{N-1}500\leq 2000$, which is satisfied for both $N=6$ and $N=2$. The results are illustrated in Fig.~\ref{fig:synthetic}(b). We observe that although the identifiability condition in \eqref{eq:common_subspace_condition3} is satisfied in both cases where 2 and 6 views are utilized, the algorithms perform better in the 6-view implementation. From both experiments we can also deduce that the proposed \texttt{RACING} works similarly to the \texttt{MAXVAR} solution and significantly outperforms \texttt{CSR-BCD}. This is a notable, considering that both \texttt{MAXVAR} and \texttt{CSR-BCD} are optimization approaches and are expected to perform better in the presence of noise.

\begin{figure}
	\centering
	\begin{subfigure}[b]{0.23\textwidth}
		\includegraphics[width=\textwidth]{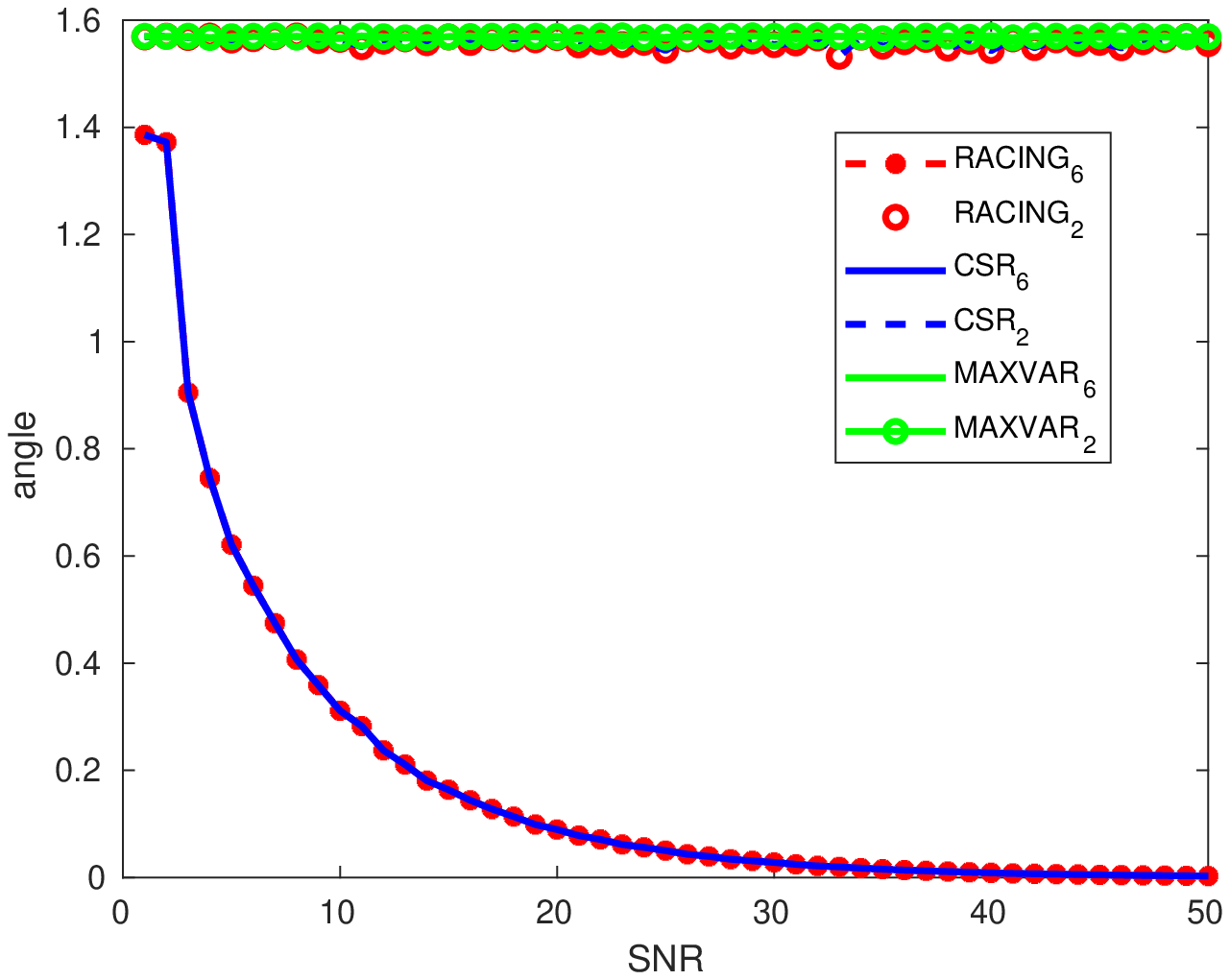}
		\caption{First scenario with approximately low-rank views}
		\label{fig:firstb}
	\end{subfigure}
	%	~ %add desired spacing between images, e. g. ~, \quad, \qquad, \hfill etc. 
	%	%(or a blank line to force the subfigure onto a new line)
	\begin{subfigure}[b]{0.23\textwidth}
		\includegraphics[width=\textwidth]{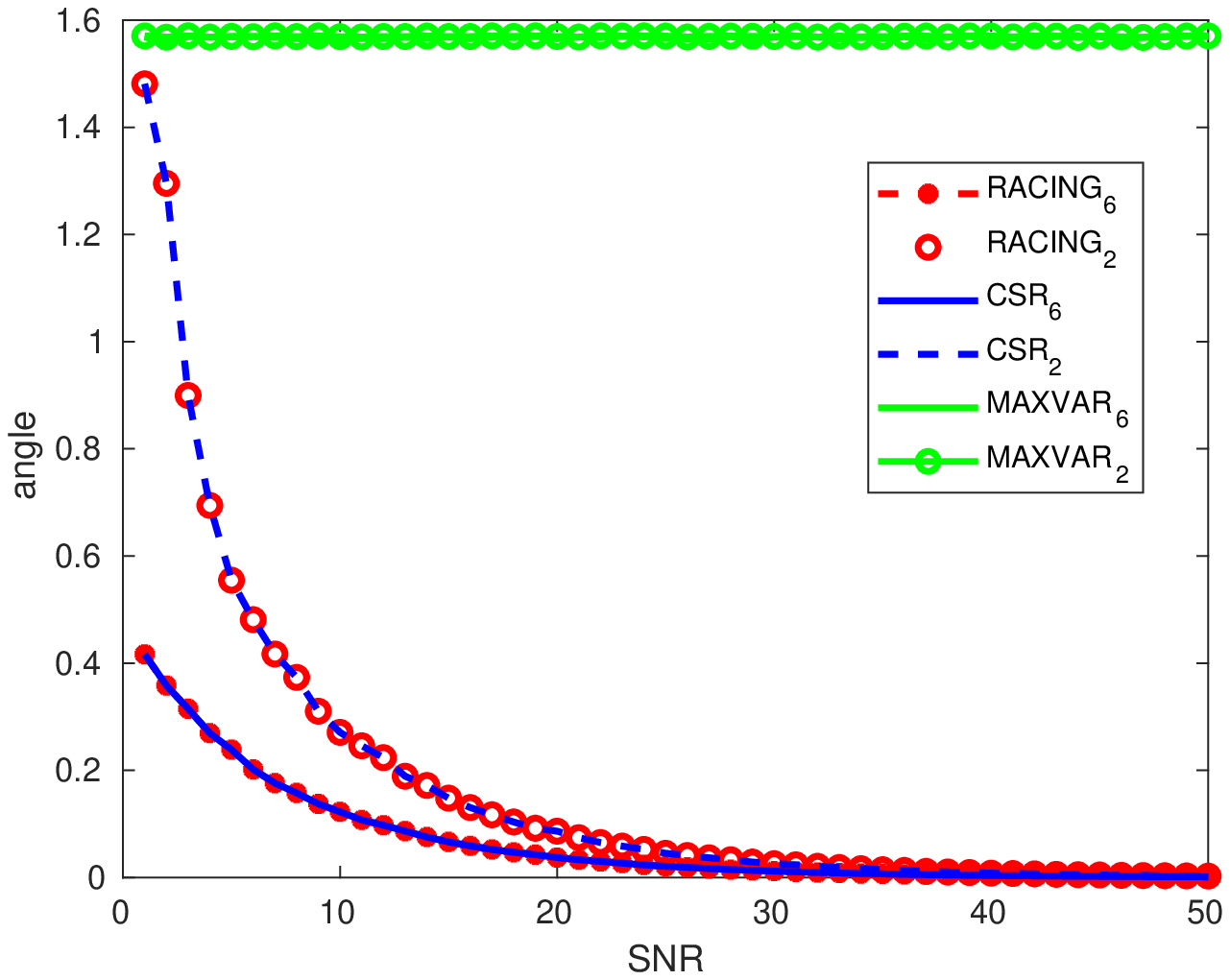}
		\caption{Second scenario with approximately low-rank views}
		\label{fig:secondb}
	\end{subfigure}
	\caption{Angle between true and recovered subspace.}
	\label{fig:synthetic2}
\end{figure}

Next we test the performance of the proposed approach and the baselines in the case where the views are approximately low rank. To this end, we generate ${\bm M},~\mymat{C}^{(n)}$ as before in scenario 1 and 2 ($I=2000,~L=1000,~R=50,~K=1050$ and $I=2000,~L=500,~R=50,~K=550$ respectively), but this time we allow $\mymat{X}^{(n)}$ to have low rank by letting $\mymat{S}^{(n)}$ to be `fat matrices', i.e., $\mymat{S}^{(n)} \in \mathbb{C}^{K \times I}$. Note that we add noise as before, so the views are technically full rank, but when the noise is small they are approximately low-rank -- i.e., they can be well-approximated by low-rank matrices. As mentioned earlier although our identifiability analysis used the full rank property of the views for simplicity, this is not in general necessary for identifiability and the subspace intersection framework. This is an important point, since although real data are typically full rank due to noise and measurement errors, the useful signal rank is often lower, and the remaining components are mostly noise. The results are presented in Fig. \ref{fig:synthetic2}.

It is clear from Fig. \ref{fig:synthetic2} that approximately low-rank views do not affect the performance of \texttt{CSR} and the proposed \texttt{RACING}. However, \texttt{MAXVAR} formulations fail to identify the common subspace. This can be explained from the fact that \texttt{MAXVAR} analysis and algorithm assume that the views are effectively full rank as mentioned earlier. On the contrary, the proposed \texttt{RACING} is allows prescribing the useful signal rank of each view, as does \texttt{CSR}. Overall, in our experiments \texttt{RACING} works better than \texttt{CSR} and comparably to \texttt{MAXVAR} in the full-rank scenarios, whereas in the low-rank scenarios it markedly outperforms \texttt{MAXVAR} and works similarly to \texttt{CSR}.

\subsection{Cross Language information Retrieval}
Finally, we test the proposed approach on the task of cross language information retrieval (CLIT). Given a set of sentences along with their translations in multiple languages the goal is to learn a low-dimensional subspace where the sentences and their translations are maximally correlated. Then, new high-dimensional sentences are projected onto the associated lower dimensional space in order to determine their translation through a database of possible choices. Note that CLIT allows fast query and search across languages, which is beneficial to machine translation systems \cite{ballesteros1997phrasal,nie1999cross,zou2013bilingual}. 

\noindent
\textbf{Data}: The dataset employed is the Europarl parallel corpus \cite{koehn2005europarl}. It contains a collection of sentences translated in 21 European languages: Romanic (French, Italian, Spanish, Portuguese, Romanian), Germanic (English, Dutch, German, Danish, Swedish), Slavik (Bulgarian, Czech, Polish, Slovak, Slovene), Finni-Ugric (Finnish, Hungarian, Estonian), Baltic (Latvian, Lithuanian), and Greek. In our experiments we only focus on the Germanic languages, i.e., English, Dutch, German, Danish and Swedish.
Each sentence is represented by a high-dimensional `bag of words' vector with $K_n=524,288$ features.

%Europarl provides multilingual alignment in the level of sentences. Each language presents a view, $\bm X^{(n)}$, and $\bm X^{(n)}(\ell,:)$ corresponds to the $\ell$th sentence in the $i$th language.

\noindent
\textbf{Procedure}: We choose a training and testing set of $I=134,227$ and $I_t=38,352$ sentences in each language-view respectively. In the training phase, the GCCA algorithms are applied to the training set in order to learn $\bm Q^{(n)},~n=1,\ldots,N$ that project each sentence to a common low dimensional subspace. In the testing phase, the testing sentences across all languages are projected onto the lower dimensional space after multiplication with $\bm Q^{(n)},~n=1,\ldots,N$. The CLIT task is completed by matching the query sentences with their translations, according to their distances in the low-dimensional subspace. 

We consider 2 scenarios. In the first one, training and testing are performed using only $N=2$ languages, i.e., Dutch and Danish. In the second scenario, the training and testing phase takes into account all Germanic languages (Dutch, Danish, English, German, Swedish).

\noindent
{\bf Evaluation}: The baseline algorithms used for comparison are \texttt{MVLSA} \cite{rastogi2015multiview}, that approximately solves the MAXVAR criterion and \texttt{GCCA-PDD} \cite{Kanatsoulis19}, which is a primal-dual algorithm that tackles the SUMCOR formulation for sparse large-scale data. We initialize \texttt{GCCA-PDD}, with \texttt{RACING} and ran for 25 iterations (total number of 5 inner and 5 outer iterations). 

To measure the performance of the competing algorithms we use the AROC and NN-freq. metric as defined in \cite{Kanatsoulis19}. The first is a ranking metric for the position of the correct translation, whereas NN-freq. measures the probability of the correct translation to be in the first place.
\noindent\\
{\bf Results}: Table \ref{tablev2gg} shows the performance of the competing algorithms for the two scenarios of CLIT. The dimension of the subspace where each sentence is projected is $R=70$ and $L_n=230$ for all views. Regarding \texttt{GCCA-PDD}, the $\rho$ parameter is set equal to 2. 
One can see that the CLIT task significantly benefits by incorporating multiple languages. To be more precise, both metrics show performance improvement when all Germanic languages are employed. For example \texttt{RACING} achieves an improvement of approximately $10\%$ in NN freq. and $1.2\%$ in AROC, when all Germanic languages are used. Note that the difference in AROC is also significant since for $N=5$ the right translation ranks in the top $290$ position, whereas for $N=2$ only in the top $520$ position. Furthermore, we observe that $PDD-GCCA$ initialized by \texttt{RACING} outperforms the other two methods, whereas \texttt{RACING} works better than \texttt{MVLSA}.
\begin{table}[H]
	\centering
	\caption{Average AROC and NN freq. (both in \%) of the Danish-Dutch CLIT.}
	\label{tablev2gg}
	\resizebox{1\linewidth}{!}{
		\begin{tabular}{ |c| c | c |c|}
			\hline
			& \multirow{2}{*}{metric}&\multicolumn{2}{| c| } {\text{\# of languages}}  \\ \cline{3-4}		
			\textbf{Algorithm}& &	\text{Danish-Dutch} & \text{5 Germanic languages} \\ \hline
			
			\multirow{2}{*}{\texttt{PDD-GCCA}}& \text{AROC}& 97.36 & 98.56 \\ \cline{2-4}
			& \text{NN freq.}& 46.34 & 69.05\\ 
			\hline
			\multirow{2}{*}{\texttt{RACING}}& \text{AROC}& 97.36 & 98.49 \\ \cline{2-4}
			& \text{NN freq.}& 46.34 & 56.97\\ 
			\hline
			
			\multirow{2}{*}{\texttt{MVLSA}}& \text{AROC}& 96.20 & 97.06 \\ \cline{2-4}
			& \text{NN freq.}& 36.72 & 41.15\\ 
			\hline

		\end{tabular}
	}
\end{table}

\section{Conclusion}
In this paper we studied generalized CCA from a linear algebraic perspective. In particular, we showed that GCCA can be interpreted as subspace intersection and provided identifiability conditions for recovering the common subspace between the views, which are relaxed compared to the standard two-view CCA. We also developed a range subspace intersection algorithm to perform GCCA, which can also handle large and high-dimensional datasets. Numerical experiments demonstrated the effectiveness of the proposed approach in the context of multi-view learning. 

\appendix
\section{Proof of equation \eqref{eq:X_MCn}}\label{app:prop}
In order to prove equation \eqref{eq:X_MCn} we first need to prove that the following properties hold without loss of generality:\\
\\
\textbf{Property 1: $\mymat{M}$, $\mymat{C}^{(1)},\ldots, \mymat{C}^{(N)}$ have full column rank.}
Without loss of generality  we can    assume that the matrices  $\mymat{M} \in \mathbb{C}^{I \times R}, \mymat{C}^{(1)} \in \mathbb{C}^{I \times L_1}, \ldots,  \mymat{C}^{(N)} \in \mathbb{C}^{I \times L_N}$ in    \eqref{eq:MFA_model} all have full column rank, Indeed, if the columns of $\mymat{M} $ are linearly dependent, then  $\mymat{M} $ can replaced by any  subset of  its columns that form a basis for $\text{range}(\mymat{M})$  {and the matrix ${\bf S}^{(n)}$ can be adjusted accordingly, without changing ${\bf X}^{(n)}$}. (Similarly for $ \mymat{C}^{(1)}, \ldots,  \mymat{C}^{(N)}$).\\
\\
\textbf{Property 2: $[\bm{M}, \bm{C}^{(1)}], \ldots, [\bm{M}, \bm{C}^{(N)}]$ have full column rank.}
\label{subsec:Property2}
First, note that the column dimension ($R+L_n$) of $[\mymat{M},\mymat{C}^{(n)}]$   {should not}  exceed its row dimension $I$, i.e., $I \geq R+L_n$. Indeed, if $I < R+L_n$, then $R+L_n-I$ columns of $ \mymat{C}^{(n)}$ can be written as linear combinations of the other $I$ columns in $[\mymat{M},\mymat{C}^{(n)}]$. Therefore these $R+L_n-I$ columns of $\mymat{C}^{(n)}$ could be discarded, while accordingly adjusting matrix ${\bf S}^{(n)}$, without changing ${\bf X}^{(n)}$.         Furthermore, we can w.l.o.g. assume that $ \text{range}( \mymat{M}) \cap \text{range}(\mymat{C}^{(n)}) =\{\myvec{0}\}$,  $n \in \{1, \ldots, N\}$, i.e., we assume w.l.o.g. that $\mymat{c}_q^{(n)} \notin \text{range}(\mymat{M})$, $q \in \{1, \ldots, L_n\}$,  $n \in \{1, \ldots, N\}$.  Indeed, if $ \mymat{c}_t^{(n)}= \mymat{M} \pmb{\beta}$ for some  $ \pmb{\beta} \in \mathbb{C}^R$, then 
\begin{align}
\mymat{X}^{(n)}=\sum_{r=1}^R \mymat{m}_r\mymat{s}_r^{(n)T}+\sum_{q=1}^{L_n} \mymat{c}_q^{(n)}\mymat{s}_r^{(n)T}=\nonumber\\\sum_{r=1}^R \mymat{m}_r(\mymat{s}_r^{(n)T}+\beta_{r}\mymat{s}_{R+t}^{(n)T})+\sum_{\substack{q=1\\ q\neq t}}^{L_n} \mymat{c}_q^{(n)}\mymat{s}_r^{(n)T}. 
\label{eq:Xn_alternative}
\end{align}
In other words, if  $  \mymat{c}_q^{(n)}  \in \text{range}(\mymat{M})$,  then we can simply consider a factorization of $  \mymat{X}^{(n)}$, as in   \eqref{eq:Xn_alternative}, that only involves a smaller $I$-by-$(L_n-1)$ matrix  $ \mymat{C}^{(n)}$. Now since $\text{range}( \mymat{M}) \cap \text{range}(\mymat{C}^{(n)}) =\{\myvec{0}\}$,  $n \in \{1, \ldots, N\}$ and $\mymat{M},~\mymat{C}^{(n)}$ have full column rank, we conclude that w.l.o.g. $[\mymat{M}, \mymat{C}^{(n)}]$ has full column rank. Then relation \eqref{eq:X_MCn} follows naturally from the fact that matrix $[\mymat{M}, \mymat{C}^{(n)}]$  has full column rank and that the subspaces    $\text{range}(\mymat{M})$ and $ \text{range}( \mymat{C}^{(n)})$ are complementary. 

\bibliography{GCCA_Bib,refs_cca_hari,refs_cca,refs_aug2015}
\bibliographystyle{ieeetr}
\onecolumn
\begin{center}
	\vspace{15pt}
	\textbf{\large Supplementary Material}
	\vspace{15pt}
\end{center}

In this section we provide an illustrative example of matrix $\pmb{\Gamma}^{(N)}$ that has full column rank for an interesting choice of $\{R, L_1, \ldots, L_N\}$. This implies that if $\mymat{M},\mymat{C}^{(n)},\mymat{S}^{(n)}, n=1,\dots,N$ are generic, then for this specific tuple that satisfies the equations in (19), $\pmb{\Gamma}^{(N)}$ is full column rank, $\text{dim}\left(\bigcap_{n=1}^N \col{\mymat{X}^{(n)}} \right)=R$ and ~$\bigcap_{n=1}^N \col{\mymat{X}^{(n)}}=\col {\mymat{M}}$. This follows our analysis in section 4. In other words, given that for this specific choice of $\{R, L_1, \ldots, L_N\}$ we can find an example of $\pmb{\Gamma}^{(N)}$ that has full column rank, det($\pmb{\Gamma}^{(N)}$) is nontrivial and any generic set of matrices $\mymat{M},\mymat{C}^{(n)},\mymat{S}^{(n)}, n=1,\dots,N$ produces a matrix $\pmb{\Gamma}^{(N)}$ that has full column rank with probability 1.

Consider the special case where $I=22$, $R=10$,  $N=3$ and $L:=L_1=L_2=L_3$. Since $2I \geq 2R+3L$, we know that $L\leq \lfloor  \frac{2}{3} (I-R)\rfloor=8$ is necessary. Hence, for the special case where  $I=22$, $R=10$,  $N=3$ and $L:=L_1=L_2=L_3=8 $
it suffices to find  a single set  of    matrices $ \mymat{M}\in \mathbb{C}^{I \times R}$,  $ \mymat{C}^{(1)} \in \mathbb{C}^{I \times L},  \mymat{C}^{(2)}\in \mathbb{C}^{I \times L}$,  $\mymat{C}^{(3)} \in \mathbb{C}^{I \times L}$   in which $I=22$, $R=10$ and $L=8$   such that the matrix
\begin{equation}
\pmb{\Gamma}^{(3)}=
\left[\begin{array}{ccccc}
\mymat{C}^{(1)} & -\mymat{M} &-\mymat{C}^{(2)} & \mymat{0}_{I,R} &   \mymat{0}_{I,L} \\
\mymat{C}^{(1)}& \mymat{0}_{I,R} &  \mymat{0}_{I,L} & -\mymat{M} &-\mymat{C}^{(3)} 
\end{array}\right]
\end{equation}
has full column rank.

The overall idea is to set $[\mymat{C}^{(1)} ,-\mymat{M} ]$ equal to the first $L+R$ columns of the $I\times I$ identity matrix $\mymat{I}_{I}$, i.e., $[\mymat{C}^{(1)} ,-\mymat{M} ]=\mymat{I}_{I}(:,1:L+R)$ and then find  appropriate (zero-one) matrices $-\mymat{C}^{(2)} $ and   $-\mymat{C}^{(3)} $ such that  $\pmb{\Gamma}^{(3)}$ has full column rank. To be more precise let $\mymat{C}^{(1)}=\mymat{I}_{I}(:,1:L)$, $\mymat{M}=\begin{bmatrix}
\mymat{0}_{I-R-L/2 \times R}\\
\mymat{I}_{R \times R}\\
\mymat{0}_{L/2 \times R}
\end{bmatrix}$, $\mymat{C}^{(2)}=\begin{bmatrix}
\mymat{0}_{L/2,L}\\
\mymat{0}_{L/2,L/2},\mymat{I}_{L/2}\\
\mymat{0}_{I-L \times L}\\
\mymat{I}_{L/2},\mymat{0}_{L/2,L/2}
\end{bmatrix}$, $\mymat{C}^{(3)}=\begin{bmatrix}
\mymat{0}_{L/2,L/2},\mymat{I}_{L/2}\\
\mymat{0}_{R \times L}\\
\mymat{I}_{L/2},\mymat{0}_{L/2,L/2}
\end{bmatrix}$. The resulting $\pmb{\Gamma}^{(3)}$ is depicted bellow. Then by inspection of the matrix it is not hard to see that $\pmb{\Gamma}^{(3)}$ has full column rank.
\\

\quad\quad\quad\quad\quad\quad\quad\quad\quad\quad\quad$\pmb{\Gamma}^{(3)}=$
\resizebox{8cm}{!}{$\left[\begin{array}{cccccccccccccccccccccccccccccccccccccccccccc}
	1 &   &   &   &   &   &   &   &   &   &   &   &   &   &   &   &   &   &   &   &   &   &   &   &   &   &   &   &   &   &   &   &   &   &   &   &   &   &   &   &   &   &   &  \\   & 1 &   &   &   &   &   &   &   &   &   &   &   &   &   &   &   &   &   &   &   &   &   &   &   &   &   &   &   &   &   &   &   &   &   &   &   &   &   &   &   &   &   &  \\   &   & 1 &   &   &   &   &   &   &   &   &   &   &   &   &   &   &   &   &   &   &   &   &   &   &   &   &   &   &   &   &   &   &   &   &   &   &   &   &   &   &   &   &  \\   &   &   & 1 &   &   &   &   &   &   &   &   &   &   &   &   &   &   &   &   &   &   &   &   &   &   &   &   &   &   &   &   &   &   &   &   &   &   &   &   &   &   &   &  \\   &   &   &   & 1 &   &   &   &   &   &   &   &   &   &   &   &   &   &   &   &   &   & 1 &   &   &   &   &   &   &   &   &   &   &   &   &   &   &   &   &   &   &   &   &  \\   &   &   &   &   & 1 &   &   &   &   &   &   &   &   &   &   &   &   &   &   &   &   &   & 1 &   &   &   &   &   &   &   &   &   &   &   &   &   &   &   &   &   &   &   &  \\   &   &   &   &   &   & 1 &   &   &   &   &   &   &   &   &   &   &   &   &   &   &   &   &   & 1 &   &   &   &   &   &   &   &   &   &   &   &   &   &   &   &   &   &   &  \\   &   &   &   &   &   &   & 1 &   &   &   &   &   &   &   &   &   &   &   &   &   &   &   &   &   & 1 &   &   &   &   &   &   &   &   &   &   &   &   &   &   &   &   &   &  \\   &   &   &   &   &   &   &   & 1 &   &   &   &   &   &   &   &   &   &   &   &   &   &   &   &   &   &   &   &   &   &   &   &   &   &   &   &   &   &   &   &   &   &   &  \\   &   &   &   &   &   &   &   &   & 1 &   &   &   &   &   &   &   &   &   &   &   &   &   &   &   &   &   &   &   &   &   &   &   &   &   &   &   &   &   &   &   &   &   &  \\   &   &   &   &   &   &   &   &   &   & 1 &   &   &   &   &   &   &   &   &   &   &   &   &   &   &   &   &   &   &   &   &   &   &   &   &   &   &   &   &   &   &   &   &  \\   &   &   &   &   &   &   &   &   &   &   & 1 &   &   &   &   &   &   &   &   &   &   &   &   &   &   &   &   &   &   &   &   &   &   &   &   &   &   &   &   &   &   &   &  \\   &   &   &   &   &   &   &   &   &   &   &   & 1 &   &   &   &   &   &   &   &   &   &   &   &   &   &   &   &   &   &   &   &   &   &   &   &   &   &   &   &   &   &   &  \\   &   &   &   &   &   &   &   &   &   &   &   &   & 1 &   &   &   &   &   &   &   &   &   &   &   &   &   &   &   &   &   &   &   &   &   &   &   &   &   &   &   &   &   &  \\   &   &   &   &   &   &   &   &   &   &   &   &   &   & 1 &   &   &   &   &   &   &   &   &   &   &   &   &   &   &   &   &   &   &   &   &   &   &   &   &   &   &   &   &  \\   &   &   &   &   &   &   &   &   &   &   &   &   &   &   & 1 &   &   &   &   &   &   &   &   &   &   &   &   &   &   &   &   &   &   &   &   &   &   &   &   &   &   &   &  \\   &   &   &   &   &   &   &   &   &   &   &   &   &   &   &   & 1 &   &   &   &   &   &   &   &   &   &   &   &   &   &   &   &   &   &   &   &   &   &   &   &   &   &   &  \\   &   &   &   &   &   &   &   &   &   &   &   &   &   &   &   &   & 1 &   &   &   &   &   &   &   &   &   &   &   &   &   &   &   &   &   &   &   &   &   &   &   &   &   &  \\   &   &   &   &   &   &   &   &   &   &   &   &   &   &   &   &   &   & 1 &   &   &   &   &   &   &   &   &   &   &   &   &   &   &   &   &   &   &   &   &   &   &   &   &  \\   &   &   &   &   &   &   &   &   &   &   &   &   &   &   &   &   &   &   & 1 &   &   &   &   &   &   &   &   &   &   &   &   &   &   &   &   &   &   &   &   &   &   &   &  \\   &   &   &   &   &   &   &   &   &   &   &   &   &   &   &   &   &   &   &   & 1 &   &   &   &   &   &   &   &   &   &   &   &   &   &   &   &   &   &   &   &   &   &   &  \\   &   &   &   &   &   &   &   &   &   &   &   &   &   &   &   &   &   &   &   &   & 1 &   &   &   &   &   &   &   &   &   &   &   &   &   &   &   &   &   &   &   &   &   &  \\ 1 &   &   &   &   &   &   &   &   &   &   &   &   &   &   &   &   &   &   &   &   &   &   &   &   &   &   &   &   &   &   &   &   &   &   &   &   &   &   &   & 1 &   &   &  \\   & 1 &   &   &   &   &   &   &   &   &   &   &   &   &   &   &   &   &   &   &   &   &   &   &   &   &   &   &   &   &   &   &   &   &   &   &   &   &   &   &   & 1 &   &  \\   &   & 1 &   &   &   &   &   &   &   &   &   &   &   &   &   &   &   &   &   &   &   &   &   &   &   &   &   &   &   &   &   &   &   &   &   &   &   &   &   &   &   & 1 &  \\   &   &   & 1 &   &   &   &   &   &   &   &   &   &   &   &   &   &   &   &   &   &   &   &   &   &   &   &   &   &   &   &   &   &   &   &   &   &   &   &   &   &   &   & 1\\   &   &   &   & 1 &   &   &   &   &   &   &   &   &   &   &   &   &   &   &   &   &   &   &   &   &   &   &   &   &   &   &   &   &   &   &   &   &   &   &   &   &   &   &  \\   &   &   &   &   & 1 &   &   &   &   &   &   &   &   &   &   &   &   &   &   &   &   &   &   &   &   &   &   &   &   &   &   &   &   &   &   &   &   &   &   &   &   &   &  \\   &   &   &   &   &   & 1 &   &   &   &   &   &   &   &   &   &   &   &   &   &   &   &   &   &   &   &   &   &   &   &   &   &   &   &   &   &   &   &   &   &   &   &   &  \\   &   &   &   &   &   &   & 1 &   &   &   &   &   &   &   &   &   &   &   &   &   &   &   &   &   &   &   &   &   &   &   &   &   &   &   &   &   &   &   &   &   &   &   &  \\   &   &   &   &   &   &   &   &   &   &   &   &   &   &   &   &   &   &   &   &   &   &   &   &   &   & 1 &   &   &   &   &   &   &   &   &   &   &   &   &   &   &   &   &  \\   &   &   &   &   &   &   &   &   &   &   &   &   &   &   &   &   &   &   &   &   &   &   &   &   &   &   & 1 &   &   &   &   &   &   &   &   &   &   &   &   &   &   &   &  \\   &   &   &   &   &   &   &   &   &   &   &   &   &   &   &   &   &   &   &   &   &   &   &   &   &   &   &   & 1 &   &   &   &   &   &   &   &   &   &   &   &   &   &   &  \\   &   &   &   &   &   &   &   &   &   &   &   &   &   &   &   &   &   &   &   &   &   &   &   &   &   &   &   &   & 1 &   &   &   &   &   &   &   &   &   &   &   &   &   &  \\   &   &   &   &   &   &   &   &   &   &   &   &   &   &   &   &   &   &   &   &   &   &   &   &   &   &   &   &   &   & 1 &   &   &   &   &   &   &   &   &   &   &   &   &  \\   &   &   &   &   &   &   &   &   &   &   &   &   &   &   &   &   &   &   &   &   &   &   &   &   &   &   &   &   &   &   & 1 &   &   &   &   &   &   &   &   &   &   &   &  \\   &   &   &   &   &   &   &   &   &   &   &   &   &   &   &   &   &   &   &   &   &   &   &   &   &   &   &   &   &   &   &   & 1 &   &   &   &   &   &   &   &   &   &   &  \\   &   &   &   &   &   &   &   &   &   &   &   &   &   &   &   &   &   &   &   &   &   &   &   &   &   &   &   &   &   &   &   &   & 1 &   &   &   &   &   &   &   &   &   &  \\   &   &   &   &   &   &   &   &   &   &   &   &   &   &   &   &   &   &   &   &   &   &   &   &   &   &   &   &   &   &   &   &   &   & 1 &   &   &   &   &   &   &   &   &  \\   &   &   &   &   &   &   &   &   &   &   &   &   &   &   &   &   &   &   &   &   &   &   &   &   &   &   &   &   &   &   &   &   &   &   & 1 &   &   &   &   &   &   &   &  \\   &   &   &   &   &   &   &   &   &   &   &   &   &   &   &   &   &   &   &   &   &   &   &   &   &   &   &   &   &   &   &   &   &   &   &   & 1 &   &   &   &   &   &   &  \\   &   &   &   &   &   &   &   &   &   &   &   &   &   &   &   &   &   &   &   &   &   &   &   &   &   &   &   &   &   &   &   &   &   &   &   &   & 1 &   &   &   &   &   &  \\   &   &   &   &   &   &   &   &   &   &   &   &   &   &   &   &   &   &   &   &   &   &   &   &   &   &   &   &   &   &   &   &   &   &   &   &   &   & 1 &   &   &   &   &  \\   &   &   &   &   &   &   &   &   &   &   &   &   &   &   &   &   &   &   &   &   &   &   &   &   &   &   &   &   &   &   &   &   &   &   &   &   &   &   & 1 &   &   &   &   \end{array}\right]$}
%\end{align}
%\appendix
%\section{Do \emph{not} have an appendix here}
%
%\textbf{\emph{Do not put content after the references.}}
%%
%Put anything that you might normally include after the references in a separate
%supplementary file.
%
%We recommend that you build supplementary material in a separate document.
%If you must create one PDF and cut it up, please be careful to use a tool that
%doesn't alter the margins, and that doesn't aggressively rewrite the PDF file.
%pdftk usually works fine. 
%
%\textbf{Please do not use Apple's preview to cut off supplementary material.} In
%previous years it has altered margins, and created headaches at the camera-ready
%stage. 
%%%%%%%%%%%%%%%%%%%%%%%%%%%%%%%%%%%%%%%%%%%%%%%%%%%%%%%%%%%%%%%%%%%%%%%%%%%%%%%
%%%%%%%%%%%%%%%%%%%%%%%%%%%%%%%%%%%%%%%%%%%%%%%%%%%%%%%%%%%%%%%%%%%%%%%%%%%%%%%

\end{document}